\documentclass[english]{lni}
\usepackage{amsmath,amsfonts}
\usepackage{algorithmic}
\usepackage{graphicx}
\usepackage{textcomp}
\usepackage{booktabs}
\usepackage{multirow}
\usepackage[table]{xcolor}
\usepackage{subcaption}
\begin{document}

\title{Classification of Unseen Identity Documents for Document Forensics
}

\author[1]{Manuel Eberhardinger}{eberhardinger@hdm-stuttgart.de}{0009-0009-2897-9250}
\author[1]{Patrick Takenaka}{takenaka@hdm-stuttgart.de}{0009-0005-2295-562X}
\author[1]{Johannes Maucher}{maucher@hdm-stuttgart.de}{0000-0002-3804-8937}
\affil[1]{Institute for Applied Artificial Intelligence, Stuttgart Media University, Stuttgart, Germany \url{https://ai.hdm-stuttgart.de} \\
}

\maketitle

\begin{abstract}
In this work, we investigate the performance of compositional generalization, i.e., the ability to combine familiar concepts in novel ways, using self-supervised visual features based on the Self-Distillation with No Labels (DINO)v2 framework. We specifically analyze the out-of-distribution performance in the case of sparse labels and apply this approach in the field of document forensics, where the necessity of a quick classification of a provided document in terms of its properties such as document type or issuing country can accelerate downstream applications including identity verification or fraud detection.

\end{abstract}

\begin{keywords}
Document Forensics \and Instance-based Classification \and  Compositional Generalization
\end{keywords}

\section{Introduction}
Document forgery is a widespread problem that causes substantial financial damage and poses challenges for security and identity verification systems. The increasing variety of identity documents and issuing authorities makes reliable authenticity assessment difficult, even for experienced document forensics experts~\cite{angel_forensic_2020}.

Document forgery develops in an adversarial setting in which issuing authorities and counterfeit producers continuously adapt to each other. New document generations introduce increasingly complex security features, such as optically variable elements, fine-grained background patterns, and specialized printing techniques. At the same time, forgery methods improve in their ability to replicate these characteristics. This leads to documents that are increasingly complex to analyze, particularly for non-expert users operating under time constraints or with limited inspection tools~\cite{angel_forensic_2020}.

In many cases, forensic analysis is performed by comparing a given document against trusted reference material. However, reference collections are often incomplete, as not all countries provide samples of their identity documents and some document types are rare or newly issued. As a result, experts rely on heterogeneous archives and visually compare document elements such as layout, printing patterns, and security features. A practical challenge in this workflow is the identification and retrieval of suitable reference documents from large collections.

Historically, this process required manual inspection of physical archives. With the availability of large digital document databases, image-based instance retrieval methods offer the possibility of automating this step and significantly reducing processing times. While longer processing times may be acceptable in forensic laboratories, other application areas such as border control or airport security require fast identification and retrieval of related documents.

Accurate identification of the document class, including both the document type and the issuing country, is an important prerequisite for downstream tasks such as class-specific document parsing and validation of document entries. Errors at this stage can negatively affect subsequent processing steps.

In this work, we study the capability of current image-based instance retrieval methods to retrieve documents of the same document type and issuing country. In addition to standard retrieval scenarios, we focus on challenging cases that are relevant for practical use. Specifically, we evaluate one-shot retrieval performance for rare document classes and zero-shot retrieval performance in a compositional generalization setting, where certain document type - country combinations are not present in the training data, while each individual attribute is still observed independently. 

We investigate these scenarios on a large, private collection of identity document images provided directly by a law enforcement agency. It contains a diverse set of document types and issuing countries and reflects real-world conditions encountered in forensic and operational settings.

We begin by discussing relevant related work in Section \ref{section:related-work}, and continue by describing our methodology and data in detail in Section \ref{section:method} and Section \ref{section:data} respectively. We show our experimental results and discussions in Section \ref{section:experiments} and conclude our work with limitations and future work in Section \ref{section:limitations}.

\section{Related Work}
\label{section:related-work}

Document image classification can rely on textual, structural, visual, or hybrid features. In this work, we focus on visual features only obtained via deep learning models, opposed to traditional models that relied on handcrafted features. This setting is especially relevant when text extraction is unreliable or costly, such as for low-quality scans or identity documents. A recent survey by Liu et al.~\cite{liu_document_2021} reviews these approaches and shows that deep visual features have become a strong alternative to text-based or handcrafted methods.

Early work in visual document classification mainly used Convolutional Neural Networks (CNNs) trained in a supervised manner. Kang et al.~\cite{kang_convolutional_2014} first demonstrated the effectiveness of CNNs for document image classification, while Harley et al.~\cite{harley_evaluation_2015} provided a systematic evaluation of CNN architectures for document classification and retrieval. To reduce the need for labeled data, Abuelwafa et al.~\cite{abuelwafa_unsupervised_2019} proposed an unsupervised CNN pretraining approach based on exemplar clustering.

More recent methods apply Vision Transformers (ViTs) \cite{dosovitskiy_image_2020} to document images. Sevim et al.~\cite{sevim_document_2022} showed that patch-based ViT models can effectively help by breaking down large images into more manageable small image areas. Li et al. \cite{li_dit_2022} further introduced the Document Image Transformer (DiT), which uses self-supervised pretraining on large collections of unlabeled document images and achieves strong performance. Related work on identity document classification focuses on supervised comparisons of different architectures under real-world constraints~\cite{voerman_optimizing_2025}.

In contrast to these approaches, our work builds on top of powerful visual representations learned from the Dino model family~\cite{caron_emerging_2021,oquab_dinov2_2023,simeoni_dinov3_2025} in a self-supervised manner. Previous work used DINOv2 for image retrieval in a neuro-symbolic anonymization pipeline \cite{eberhardinger2025anonymization} or evaluated how different anonymization methods impact the performance of DINOv2 on image retrieval \cite{chen2026evaluating}. 

To the best of our knowledge, only the work of Zenk et al.~\cite{zenk_investigations_2023} similarly builds on top of Dino~\cite{caron_emerging_2021,oquab_dinov2_2023} for classification of document properties. However, they focus on classification of local features such as font, script, and location of historical documents. In our case, we investigate more global features such as issuing country or document type of identity documents. Furthermore, our focus is on the sparse multi-label classification performance that is more relevant for document forensics.

The field of compositional generalization addresses the cognitive and computational capacity to recombine familiar visual components into novel, unseen configurations \cite{fodor1988connectionism, lake2017building, keysers2019measuring}. This is analogous to human cognition, where familiarity with subconcepts such as ``red cube'' and ``blue sphere'' inherently enables the immediate understanding of ``red sphere'' \cite{lake2017building}. This study investigates the degree to which the model abstracts the visual features e.g. of a passport. Specifically, we evaluate whether the model can successfully generalize this abstract concept to passports from previously unseen issuing countries.

\section{Method}
\label{section:method}
A multi-stage approach is employed for the classification of document images. The objective is to assign document scans to both a specific document type and its corresponding issuing country. The overall process is divided into three primary phases:

\begin{enumerate}
    \item Fine-tuning a DINOv2 model \cite{oquab_dinov2_2023} on the training dataset,
    \item Generating embeddings for all images using the trained model,
    \item Implementing a classification mechanism based on these embeddings.
\end{enumerate}

Following the training of the DINOv2 model, all images in the dataset are transformed into embeddings. These serve as compact, numerical representations of the visual content, enabling efficient downstream processing. Classification is subsequently performed using these generated embeddings. To achieve this, all training embeddings are indexed in a vector database powered by Faiss \cite{douze2025faiss}. During inference, a sample from the test set is used to query this database. We employ a $k$-Nearest Neighbors (kNN) approach \cite{aha1991instance}, where the database retrieves the documents from the training set that shows the highest embedding similarity to the query calculated via the cosine distance. Specifically, the five most similar documents are extracted, and the final classification of the test sample is determined through a majority vote of these neighbors.

\section{Data}
\label{section:data}
The full dataset consists of digitized identity documents from a wide range of issuing countries and document types. It represents a snapshot of a document database made available for this research. The data is highly heterogeneous, with a strong imbalance in the number of samples per issuing country and document type.

The dataset contains both authentic documents and forgeries. Overall, as seen in Table \ref{tab:dataset_comparison} the full dataset includes 21{,}826 documents and 238{,}389 images, with multiple images available per document. Document types are defined according to the classification scheme used by the law enforcement forensics group.

\begin{table}[t]
\centering
\small
\caption{Summary statistics of the full dataset and the VisoCore-based evaluation subset.}
\label{tab:dataset_comparison}
\begin{tabular}{lrr}
\hline
\textbf{Property} & \textbf{Full dataset} & \textbf{VisoCore subset} \\
\hline
Documents & 21{,}826 & 6{,}852 \\
Images & 238{,}389 & 38{,}327 \\
Issuing countries & 155 & 137 \\
Document types & 188 & 65 \\
Classes (type--country) & 999 & 455 \\
\hline
\end{tabular}
\end{table}

The most frequent document types are passports (33.7\%), national driver's licenses (24.4\%), and ID cards (18.7\%). The most common issuing countries are Iraq (26.6\%), Syria (13.4\%), and Afghanistan (7.4\%). The most frequent document type--country combinations include Iraqi passports (26{,}287 images), Syrian passports (18{,}264 images), and Iraqi ID cards (17{,}223 images).

Due to the strong heterogeneity of the full dataset, we construct a more controlled evaluation subset. We restrict this subset to images captured using the VisoCore~\cite{noauthor_visocore_nodate} document scanner, a device commonly deployed at border control checkpoints and government agencies. The scanner provides standardized image acquisition conditions, reducing variability caused by different capture devices.

This restriction limits the evaluation to commonly encountered identity document formats and results in a more comparable setting. The resulting subset contains 38{,}327 images corresponding to 6{,}852 documents from 137 issuing countries and 65 document types, forming 455 document type--country classes (see \Cref{tab:dataset_comparison} VisoCore subset).

\begin{table}[t]
\centering
\small
\caption{Most frequent issuing countries and document types in the VisoCore subset.}
\label{tab:visocore_distribution}
\begin{tabular}{l r @{\hspace{1em}} l r}
\hline
\textbf{Country} & \textbf{\%} & \textbf{Doc. type} & \textbf{\%} \\
\hline
Syria & 10.9 & National driver's license & 46.3 \\
Iraq & 8.5 & Passport & 23.9 \\
Poland & 6.6 & ID card & 20.6 \\
Italy & 5.5 & Residence permit & 3.9 \\
Ukraine & 5.3 & Int. driver's license & 1.3 \\
\hline
\end{tabular}
\end{table}

The most frequent document type--country combinations in this subset are Syrian passports (2{,}570 images), Iraqi passports (2{,}401 images), Polish national driver's licenses (2{,}154 images), Ukrainian national driver's licenses (1{,}566 images), and Italian national driver's licenses (1{,}266 images).

From the VisoCore-based subset, we define training and evaluation splits targeting different retrieval and generalization scenarios. The evaluation set is constructed as the union of four disjoint subsets:

\begin{itemize}
    \item \textbf{Regular evaluation:} For 5\% of the classes, one document is assigned to evaluation and the remaining documents are used for training.
    \item \textbf{Compositional generalization:} For 5\% of the classes, all documents are assigned to evaluation and none are included in training.
    \item \textbf{One-shot issuing country:} For 10\% of the issuing countries, all documents except one are assigned to evaluation, with a single document retained for training.
    \item \textbf{One-shot document type:} For 10\% of the document types, all documents except one are assigned to evaluation, with a single document retained for training.
\end{itemize}

The selected classes are non-overlapping across the four subsets. For the regular evaluation and compositional generalization settings, we ensure that at least one authentic document remains in the training set. The different subsets of the evaluation data then allow us to investigate the performance of the classification model from different perspectives later on. The training data consists of the full document dataset minus the documents selected for evaluation. 
    
\section{Experiments}
\label{section:experiments}

\subsection{Experimental Setup}
\begin{table}[th]
    \centering
    \resizebox{\linewidth}{!}{
    \begin{tabular}{lrrccc}
        \toprule
        \textbf{Name} & \textbf{No. of Parameters} & \textbf{Embedding ($d$)} & \textbf{Heads} & \textbf{Blocks} & \textbf{Architecture} \\
        \midrule
        Small & 21 Mio. & 384 & 6 & 12 & ViT-S/16 \\
        Base & 86 Mio. & 768 & 12 & 18 & ViT-B/16 \\
        Large & 300 Mio. & 1024 & 16 & 24 & ViT-L/16 \\
        Giant & 1.100 Mio. & 1536 & 24 & 40 & ViT-g/16 \\
        \bottomrule
    \end{tabular}}
    \caption{Architecture details of the different ViT-networks used in this work. Embedding ($d$) is the size of the embedding vector, Heads are the number of heads in the Multi-head attention, and Blocks represents the number of Transformer blocks used. }
    \label{tab:dinov2-specs}
\end{table}
\Cref{tab:dinov2-specs} shows architecture details of the different ViT-networks used in this work. 
Embedding ($d$) is the size of the embedding vector, Heads are the number of heads in the Multi-head attention, and Blocks represents the number of Transformer blocks used. 
All models use a patch size of 16 pixels and the fast training configuration\footnote{\url{https://github.com/facebookresearch/dinov2/blob/main/dinov2/configs/train/vitl16_short.yaml}}.
To reduce sensitivity to random selection and class imbalance during generation of the evaluation data, we generate three folds using different random seeds and report both the results for all folds and the average of each individual fold. All models are trained with the default hyperparameters reported in \cite{oquab_dinov2_2023}. We report the standard accuracy metric for classification, which is calculated by dividing the number of correct predictions by the total number of predictions. 

\subsection{Evaluation}
\begin{table}[th!]
\centering
    \small % Optimale Schriftgröße für dichte Datentabellen
    \setlength{\tabcolsep}{2pt} % Spaltenabstand kompakt halten
\resizebox{\linewidth}{!}{
    \begin{tabular}{l rr rr rr rr}
    \toprule
    & \multicolumn{2}{c}{\textbf{Small}} & \multicolumn{2}{c}{\textbf{Base}} & \multicolumn{2}{c}{\textbf{Large}} & \multicolumn{2}{c}{\textbf{Giant}} \\
    \cmidrule(lr){2-3} \cmidrule(lr){4-5} \cmidrule(lr){6-7} \cmidrule(lr){8-9}
    \textbf{Kategorie / Fold} & \text{Doc} & \text{Country} & \text{Doc} & \text{Country} & \text{Doc} & \text{Country} & \text{Doc} & \text{Country} \\
    \midrule
    \textit{Fold 1} & & & & & & & & \\
    \quad Regular & 87.1 & 78.8 & 87.1 & 83.3 & 88.6 & 82.6 & 76.5 & 60.1 \\
    \quad Comp. Gen. & 67.9 & 6.7 & 69.6 & 6.4 & 71.9 & 9.1 & 48.6 & 4.9 \\
    \quad One-Shot C. & 78.6 & 35.6 & 81.7 & 37.9 & 81.5 & 37.2 & 67.2 & 11.8 \\
    \quad One-Shot D. & 9.4 & 77.8 & 10.2 & 78.3 & 9.9 & 78.9 & 4.4 & 73.2 \\
    \quad Overall & 66.6 & 32.1 & 68.8 & 33.4 & 69.5 & 34.1 & 53.2 & 18.9 \\
    \midrule
    \textit{Fold 2} & & & & & & & & \\
    \quad Regular & 82.6 & 71.2 & 84.1 & 70.5 & 81.8 & 70.5 & 68.2 & 46.2 \\
    \quad Comp. Gen. & 73.4 & 31.4 & 78.5 & 29.6 & 81.3 & 30.4 & 66.6 & 14.3 \\
    \quad One-Shot C. & 84.1 & 49.2 & 83.2 & 48.5 & 84.6 & 51.5 & 68.7 & 19.7 \\
    \quad One-Shot D. & 49.6 & 65.9 & 48.9 & 69.6 & 40.9 & 73.9 & 28.6 & 48.9 \\
    \quad Overall & 78.2 & 43.1 & 79.8 & 42.1 & 81.3 & 44.2 & 66.1 & 19.3 \\
    \midrule
    \textit{Fold 3} & & & & & & & & \\
    \quad Regular & 98.0 & 84.6 & 96.5 & 83.6 & 94.7 & 86.4 & 84.3 & 60.4 \\
    \quad Comp. Gen. & 28.6 & 47.6 & 30.8 & 48.3 & 30.3 & 48.8 & 28.9 & 46.5 \\
    \quad One-Shot C. & 84.2 & 56.1 & 84.1 & 58.4 & 88.4 & 56.9 & 54.6 & 23.7 \\
    \quad One-Shot D. & 20.8 & 73.6 & 20.8 & 74.3 & 22.9 & 75.0 & 9.0 & 65.3 \\
    \quad Overall & 79.1 & 56.3 & 79.1 & 58.4 & 82.9 & 57.1 & 52.3 & 26.8 \\
    \midrule
    \textbf{Average} & & & & & & & & \\
    \quad Regular & $89.2 \pm 7.9$ & $78.2 \pm 6.7$ & $89.2 \pm 6.4$ & $79.1 \pm 7.5$ & $88.4 \pm 6.4$ & $79.8 \pm 8.3$ & $76.3 \pm 8.1$ & $55.7 \pm 8.2$ \\
    \quad Comp. Gen. & $56.6 \pm 24.4$ & $28.6 \pm 20.6$ & $59.6 \pm 25.4$ & $28.1 \pm 21.0$ & $61.2 \pm 27.1$ & $29.4 \pm 19.9$ & $48.0 \pm 18.8$ & $21.9 \pm 21.8$ \\
    \quad One-Shot C. & $82.3 \pm 3.2$ & $47.0 \pm 10.4$ & $83.0 \pm 1.2$ & $48.3 \pm 10.3$ & $84.8 \pm 3.5$ & $48.5 \pm 10.2$ & $63.5 \pm 7.8$ & $18.4 \pm 6.0$ \\
    \quad One-Shot D. & $26.6 \pm 20.7$ & $72.5 \pm 6.0$ & $26.6 \pm 20.0$ & $74.1 \pm 4.4$ & $24.6 \pm 15.6$ & $75.9 \pm 2.6$ & $14.0 \pm 12.8$ & $62.5 \pm 12.4$ \\
    \quad Overall & $\mathbf{74.6 \pm 7.0}$ & $\mathbf{43.9 \pm 12.1}$ & $\mathbf{75.9 \pm 6.2}$ & $\mathbf{44.6 \pm 12.7}$ & $\mathbf{77.9 \pm 7.3}$ & $\mathbf{45.1 \pm 11.6}$ & $\mathbf{57.2 \pm 7.7}$ & $\mathbf{21.7 \pm 4.5}$ \\
    \bottomrule
    \end{tabular}}
    \caption{Results of the four DINOv2 architectures on the three folds of the DOKIQ dataset for evaluating the four different scenarios. The values correspond to the classification accuracy (in percent) using the kNN classifier with $ k=5$.}
    \label{tab:eval-dokiq}
\end{table}
\cref{tab:eval-dokiq} displays the classification accuracy. A key finding is that model performance remains largely stable across the Small, Base, and Large architecture variants. The highest overall accuracy is achieved by the Large model (0.779 for \textit{Document Type}, 0.451 for \textit{Country}). However, the performance gain from Small to Large is marginal, suggesting that even the smallest architecture is capable of efficiently encoding the essential structural features of the documents. 
Notably, a significant degradation in performance is observed with the Giant model, where accuracy scores drop significantly across all categories. This decline may be attributed to an excessively large model capacity relative to the dataset size, or to the inherent challenges of kNN classification in extremely high-dimensional spaces (1,536 dimensions), where distance metrics often lose their discriminative power \cite{beyer1999when}.
Across all model architectures, document type recognition consistently outperforms country classification. 
This result is consistent with expectations: document types are typically characterized by distinct global layouts, whereas countries are often distinguished by fine-grained details such as national flags or subtle textual nuances. These finer details may be weighted less heavily during the self-supervised pre-training of DINOv2, as they are less likely to be preserved or prioritized under standard image augmentation techniques.
The substantial fluctuations observed between different folds (e.g. \textit{Compositional Generalization} in \textit{Fold 1} vs. \textit{Fold 3}) indicate that the quality of the evaluation is highly dependent on the composition of the reference database. \textit{Fold 3}, in particular, shows exceptionally high accuracy for document type recognition in the \textit{Regular} set, suggesting a more favorable distribution of training and test data within that split. 

The evaluation of Out-of-Distribution performance reveals several noteworthy trends:
\begin{itemize}
    \item \textbf{Compositional Generalization:} Performance in country recognition drops significantly across all folds. The model struggles to correctly identify the issuing country when presented with a novel combination of document type and country.
    \item \textbf{One-Shot Scenarios:} Interestingly, One-Shot Country recognition performs considerably better than One-Shot Document Type. When the model encounters a document type it has seen only once or not at all, classification accuracy remains low, even if the country is familiar. This suggests that document type information is more intrinsically tied to specific visual instances than country-specific information.
\end{itemize}

\subsection{Visualization of Document Similarity}
\begin{figure}[tbp]
    \centering
    % Left Subfigure
    \begin{subfigure}[tb]{0.45\linewidth}
        \centering
        \includegraphics[width=\linewidth]{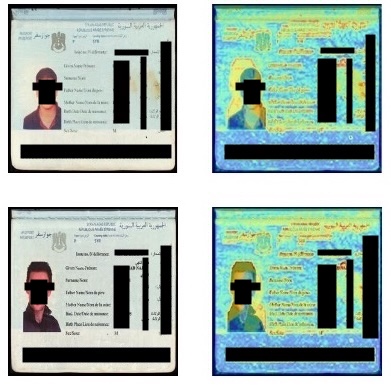}
        \caption{two Syrian passports}
        \label{fig:similarity1}
    \end{subfigure}
    \hfill % Adds horizontal space between the images
    % Right Subfigure
    \begin{subfigure}[tb]{0.45\linewidth}
        \centering
        \includegraphics[width=\linewidth]{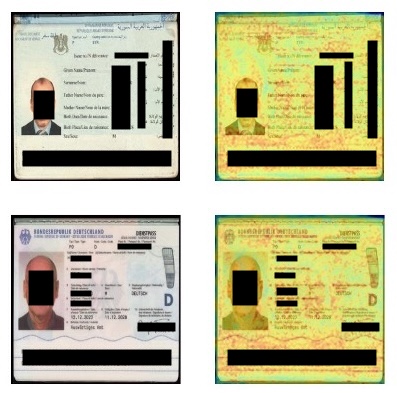}
        \caption{German passport vs. Syrian passport}
        \label{fig:similarity2}
    \end{subfigure}
    
    \caption{Visual comparison of patch-token embeddings. Blue tones indicate high semantic correspondence, while red/yellow indicate divergence.}
    \label{fig:similarity_comparison}
\end{figure}
To evaluate the quality of the learned embeddings and provide first steps into explainability of these results, we examine the local feature similarity between image pairs based on their patch tokens, i.e, the embeddings of each 16×16 patch of the divided input image \cite{dosovitskiy_image_2020}. For each pair of corresponding token positions $(i, j)$, we visualize the inverted cosine similarity ($1 - \text{similarity}$) as a heatmap over both documents. Cool tones (blue) represent high semantic correspondence, while warm tones (yellow/red) indicate significant differences. The heatmaps in \cref{fig:similarity_comparison} verify the spatial consistency of the DINOv2 representations. For documents belonging to the same category, the dominant blue coloration confirms that the model extracts stable, spatially consistent structural features across different instances. Conversely, the high divergence in comparisons of distinct document classes highlights the model’s discriminative power. From an explainability perspective, these visualizations serve as a post-hoc verification that the model focuses on relevant document structures rather than background noise or dataset artifacts, providing a transparent justification for its classification decisions. Building on these findings, more sophisticated explainability methods could be developed to support border control experts, e.g., by generating counterfactual explanations \cite{molnar2025} to highlight missing or deviating security features that differentiate authentic documents from fraudulent ones.

\section{Limitations and Future Work}
\label{section:limitations}

While our approach demonstrates the efficacy of using self-supervised embeddings for document classification, several limitations remain that leave opportunities for future research.
A primary limitation of the current study is the granularity of the classification. Our analysis was confined to two categories: document type and issuing country. We did not investigate specific document models or versions within these categories. In real-world scenarios, distinguishing between different generations of the same document type is important, as security features and the layout of documents often evolve. 

Additionally, the reliance on purely visual embeddings means that textual information, which could be used for distinguishing between similar-looking countries, is only implicitly captured. The observed difficulty in country identification compared to document type recognition suggests that the self-supervised pre-training may prioritize global structural features over the fine-grained textual details necessary for identifying the issuing country. 
A multimodal approach that combines visual embeddings from DINOv2 with data from optical character recognition (OCR) could mitigate this problem by incorporating textual cues (e.g. country names or specific keywords) when the visual layouts of documents are nearly identical.

\paragraph{Acknowledgments}
We thank the Bundesdruckerei GmbH for funding this research project. We extend our gratitude to the Document Forensics Department of the Baden-Württemberg State Office of Criminal Investigations for their invaluable assistance in providing the printed documents essential to this project.

\paragraph{Declaration on Generative AI}
During the preparation of this work, the author(s) used ChatGPT to improve formulation, grammar, and readability. After using this tool/service, the author(s) reviewed and edited the content as needed and take full responsibility for the publication’s content.

\printbibliography

@inproceedings{dosovitskiy_image_2020,
	title = {An Image is Worth 16x16 Words: Transformers for Image Recognition at Scale},
	shorttitle = {An Image is Worth 16x16 Words},
	abstract = {While the Transformer architecture has become the de-facto standard for natural language processing tasks, its applications to computer vision remain limited. In vision, attention is either applied in conjunction with convolutional networks, or used to replace certain components of convolutional networks while keeping their overall structure in place. We show that this reliance on {CNNs} is not necessary and a pure transformer applied directly to sequences of image patches can perform very well on image classification tasks. When pre-trained on large amounts of data and transferred to multiple mid-sized or small image recognition benchmarks ({ImageNet}, {CIFAR}-100, {VTAB}, etc.), Vision Transformer ({ViT}) attains excellent results compared to state-of-the-art convolutional networks while requiring substantially fewer computational resources to train.},
	eventtitle = {International Conference on Learning Representations},
	author = {Dosovitskiy, Alexey and Beyer, Lucas and Kolesnikov, Alexander and Weissenborn, Dirk and Zhai, Xiaohua and Unterthiner, Thomas and Dehghani, Mostafa and Minderer, Matthias and Heigold, Georg and Gelly, Sylvain and Uszkoreit, Jakob and Houlsby, Neil},
	%urldate = {2026-02-09},
	date = {2020-10-02},
	langid = {english},
}

@inproceedings{kang_convolutional_2014,
	title = {Convolutional Neural Networks for Document Image Classification},
	issn = {1051-4651},
	doi = {10.1109/ICPR.2014.546},
	abstract = {This paper presents a Convolutional Neural Network ({CNN}) for document image classification. In particular, document image classes are defined by the structural similarity. Previous approaches rely on hand-crafted features for capturing structural information. In contrast, we propose to learn features from raw image pixels using {CNN}. The use of {CNN} is motivated by the the hierarchical nature of document layout. Equipped with rectified linear units and trained with dropout, our {CNN} performs well even when document layouts present large inner-class variations. Experiments on public challenging datasets demonstrate the effectiveness of the proposed approach.},
	pages = {3168--3172},
	booktitle = {2014 22nd International Conference on Pattern Recognition},
	author = {Kang, Le and Kumar, Jayant and Ye, Peng and Li, Yi and Doermann, David},
	%urldate = {2026-02-09},
	date = {2014-08},
	%note = {{ISSN}: 1051-4651},
	keywords = {Accuracy, Computer vision, Kernel, Layout, {NIST}, Pattern recognition, Training},
}

@article{oquab_dinov2_2023,
	title = {{DINOv}2: Learning Robust Visual Features without Supervision},
	issn = {2835-8856},
	shorttitle = {{DINOv}2},
	journaltitle = {Transactions on Machine Learning Research},
	author = {Oquab, Maxime and Darcet, Timothée and Moutakanni, Théo and Vo, Huy V. and Szafraniec, Marc and Khalidov, Vasil and Fernandez, Pierre and Haziza, Daniel and Massa, Francisco and El-Nouby, Alaaeldin and Assran, Mido and Ballas, Nicolas and Galuba, Wojciech and Howes, Russell and Huang, Po-Yao and Li, Shang-Wen and Misra, Ishan and Rabbat, Michael and Sharma, Vasu and Synnaeve, Gabriel and Xu, Hu and Jegou, Herve and Mairal, Julien and Labatut, Patrick and Joulin, Armand and Bojanowski, Piotr},
	urldate = {2026-02-09},
	date = {2023-07-26},
	langid = {english},
}

@article{douze2025faiss,
  title={The faiss library},
  author={Douze, Matthijs and Guzhva, Alexandr and Deng, Chengqi and Johnson, Jeff and Szilvasy, Gergely and Mazar{\'e}, Pierre-Emmanuel and Lomeli, Maria and Hosseini, Lucas and J{\'e}gou, Herv{\'e}},
  journal={IEEE Transactions on Big Data},
  year={2025},
  publisher={IEEE}
}

@misc{simeoni_dinov3_2025,
	title = {{DINOv}3},
	doi = {10.48550/arXiv.2508.10104},
	number = {{arXiv}:2508.10104},
	publisher = {{arXiv}},
	author = {Siméoni, Oriane and Vo, Huy V. and Seitzer, Maximilian and Baldassarre, Federico and Oquab, Maxime and Jose, Cijo and Khalidov, Vasil and Szafraniec, Marc and Yi, Seungeun and Ramamonjisoa, Michaël and Massa, Francisco and Haziza, Daniel and Wehrstedt, Luca and Wang, Jianyuan and Darcet, Timothée and Moutakanni, Théo and Sentana, Leonel and Roberts, Claire and Vedaldi, Andrea and Tolan, Jamie and Brandt, John and Couprie, Camille and Mairal, Julien and Jégou, Hervé and Labatut, Patrick and Bojanowski, Piotr},
	urldate = {2026-02-10},
	date = {2025-08-13},
	eprinttype = {arxiv},
	eprint = {2508.10104 [cs]},
}

@inproceedings{li_dit_2022,
	location = {New York, {NY}, {USA}},
	title = {{DiT}: Self-supervised Pre-training for Document Image Transformer},
	isbn = {978-1-4503-9203-7},
	doi = {10.1145/3503161.3547911},
	series = {{MM} '22},
	shorttitle = {{DiT}},
	abstract = {Image Transformer has recently achieved significant progress for natural image understanding, either using supervised ({ViT}, {DeiT}, etc.) or self-supervised ({BEiT}, {MAE}, etc.) pre-training techniques. In this paper, we propose {DiT}, a self-supervised pre-trained Document Image Transformer model using large-scale unlabeled text images for Document {AI} tasks, which is essential since no supervised counterparts ever exist due to the lack of human-labeled document images. We leverage {DiT} as the backbone network in a variety of vision-based Document {AI} tasks, including document image classification, document layout analysis, table detection as well as text detection for {OCR}. Experiment results have illustrated that the self-supervised pre-trained {DiT} model achieves new state-of-the-art results on these downstream tasks, e.g. document image classification (91.11 - 92.69), document layout analysis (91.0 - 94.9), table detection (94.23 - 96.55) and text detection for {OCR} (93.07 - 94.29). The code and pre-trained models are publicly available at https://aka.ms/msdit.},
	pages = {3530--3539},
	booktitle = {Proceedings of the 30th {ACM} International Conference on Multimedia},
	publisher = {Association for Computing Machinery},
	author = {Li, Junlong and Xu, Yiheng and Lv, Tengchao and Cui, Lei and Zhang, Cha and Wei, Furu},
	%urldate = {2026-02-09},
	date = {2022},
}

@article{lake2017building,
  title={Building machines that learn and think like people},
  author={Lake, Brenden M and Ullman, Tomer D and Tenenbaum, Joshua B and Gershman, Samuel J},
  journal={Behavioral and brain sciences},
  volume={40},
  pages={253},
  year={2017},
  publisher={Cambridge University Press}
}

@article{fodor1988connectionism,
    title = {Connectionism and cognitive architecture: A critical analysis},
    journal = {Cognition},
    volume = {28},
    number = {1},
    pages = {3-71},
    year = {1988},
    issn = {0010-0277},
    doi = {https://doi.org/10.1016/0010-0277(88)90031-5},
    %url = {https://www.sciencedirect.com/science/article/pii/0010027788900315},
    author = {Jerry A. Fodor and Zenon W. Pylyshyn},
}

@inproceedings{
    keysers2019measuring,
    title={Measuring Compositional Generalization: A Comprehensive Method on Realistic Data},
    author={Daniel Keysers and Nathanael Sch{\"a}rli and Nathan Scales and Hylke Buisman and Daniel Furrer and Sergii Kashubin and Nikola Momchev and Danila Sinopalnikov and Lukasz Stafiniak and Tibor Tihon and Dmitry Tsarkov and Xiao Wang and Marc van Zee and Olivier Bousquet},
    booktitle={International Conference on Learning Representations},
    year={2020},
    %url={https://openreview.net/forum?id=SygcCnNKwr}
}

@book{molnar2025,
  title={Interpretable Machine Learning},
  subtitle={A Guide for Making Black Box Models Explainable},
  author={Christoph Molnar},
  year={2025},
  edition={3},
  isbn={978-3-911578-03-5},
  url={https://christophm.github.io/interpretable-ml-book}
}

@article{aha1991instance,
  title={Instance-based learning algorithms},
  author={Aha, David W and Kibler, Dennis and Albert, Marc K},
  journal={Machine learning},
  volume={6},
  number={1},
  pages={37--66},
  year={1991},
  publisher={Springer}
}

@inproceedings{eberhardinger2025anonymization,
  title={Anonymization of documents for law enforcement with machine learning},
  author={Eberhardinger, Manuel and Takenaka, Patrick and Grie{\ss}haber, Daniel and Maucher, Johannes},
  booktitle={2025 IEEE Symposium on Computational Intelligence in Security, Defence and Biometrics (CISDB)},
  pages={1--7},
  year={2025},
  organization={IEEE}
}

@article{chen2026evaluating,
  title={Evaluating the Impact of Data Anonymization on Image Retrieval},
  author={Chen, Marvin and Eberhardinger, Manuel and Maucher, Johannes},
  journal={arXiv preprint arXiv:2602.19641},
  year={2026}
}

@inproceedings{beyer1999when,
    author="Beyer, Kevin
    and Goldstein, Jonathan
    and Ramakrishnan, Raghu
    and Shaft, Uri",
    editor="Beeri, Catriel
    and Buneman, Peter",
    title="When Is ``Nearest Neighbor'' Meaningful?",
    booktitle="Database Theory --- ICDT'99",
    year="1999",
    publisher="Springer Berlin Heidelberg",
    address="Berlin, Heidelberg",
    pages="217--235",
    isbn="978-3-540-49257-3"
}

@article{liu_document_2021,
	title = {Document image classification: Progress over two decades},
	volume = {453},
	issn = {0925-2312},
	doi = {10.1016/j.neucom.2021.04.114},
	shorttitle = {Document image classification},
	pages = {223--240},
	journaltitle = {Neurocomputing},
	shortjournal = {Neurocomputing},
	author = {Liu, Li and Wang, Zhiyu and Qiu, Taorong and Chen, Qiu and Lu, Yue and Suen, Ching Y.},
	urldate = {2026-02-09},
	date = {2021-09-17},
}

@inproceedings{zenk_investigations_2023,
	%location = {New York, {NY}, {USA}},
	title = {Investigations on Self-supervised Learning for Script-, Font-type, and Location Classification on Historical Documents},
	isbn = {979-8-4007-0841-1},
	doi = {10.1145/3604951.3605519},
	series = {{HIP} '23},
	abstract = {In the context of automated classification of historical documents, we investigate three contemporary self-supervised learning ({SSL}) techniques ({SimSiam}, Dino, and {VICReg}) for the pre-training of three different document analysis tasks, namely script-type, font-type, and location classification. Our study draws samples from multiple datasets that contain images of manuscripts, prints, charters, and letters. The representations derived via pre-text training are taken as inputs for k-{NN} classification and a parametric linear classifier. The latter is placed atop the pre-trained backbones to enable fine-tuning of the entire network to further improve the classification by exploiting task-specific label data. The network’s final performance is assessed via independent test sets obtained from the {ICDAR}2021 Competition on Historical Document Classification. We empirically show that representations learned with {SSL} are significantly better suited for subsequent document classification than features generated by commonly used transfer learning on {ImageNet}.},
	%pages = {97--102},
	booktitle = {Proceedings of the 7th International Workshop on Historical Document Imaging and Processing},
	%publisher = {Association for Computing Machinery},
	author = {Zenk, Johan and Kordon, Florian and Mayr, Martin and Seuret, Mathias and Christlein, Vincent},
	%urldate = {2026-02-09},
	date = {2023-08-25},
}

@article{voerman_optimizing_2025,
	title = {Optimizing identity documents classification in online systems: A comparative analysis},
	issn = {1433-2825},
	doi = {10.1007/s10032-025-00555-5},
	shorttitle = {Optimizing identity documents classification in online systems},
	journaltitle = {International Journal on Document Analysis and Recognition ({IJDAR})},
	shortjournal = {{IJDAR}},
	author = {Voerman, Joris and Al-Ghadi, Musab and Sidere, Nicolas and Coustaty, Mickael and Lessard, Olivier},
	urldate = {2026-02-09},
	date = {2025-09-08},
	langid = {english},
}

@article{abuelwafa_unsupervised_2019,
	title = {Unsupervised Exemplar-Based Learning for Improved Document Image Classification},
	volume = {7},
	issn = {2169-3536},
	doi = {10.1109/ACCESS.2019.2940884},
	pages = {133738--133748},
	journaltitle = {{IEEE} Access},
	author = {Abuelwafa, Sherif and Pedersoli, Marco and Cheriet, Mohamed},
	urldate = {2026-02-09},
	date = {2019},
}

@inproceedings{harley_evaluation_2015,
	title = {Evaluation of deep convolutional nets for document image classification and retrieval},
	doi = {10.1109/ICDAR.2015.7333910},
	abstract = {This paper presents a new state-of-the-art for document image classification and retrieval, using features learned by deep convolutional neural networks ({CNNs}). In object and scene analysis, deep neural nets are capable of learning a hierarchical chain of abstraction from pixel inputs to concise and descriptive representations. The current work explores this capacity in the realm of document analysis, and confirms that this representation strategy is superior to a variety of popular handcrafted alternatives. Extensive experiments show that (i) features extracted from {CNNs} are robust to compression, (ii) {CNNs} trained on non-document images transfer well to document analysis tasks, and (iii) enforcing region-specific feature-learning is unnecessary given sufficient training data. This work also makes available a new labelled subset of the {IIT}-{CDIP} collection, containing 400,000 document images across 16 categories.},
	pages = {991--995},
	booktitle = {2015 13th International Conference on Document Analysis and Recognition ({ICDAR})},
	author = {Harley, Adam W. and Ufkes, Alex and Derpanis, Konstantinos G.},
	%urldate = {2026-02-09},
	date = {2015-08},
	keywords = {Image coding, Libraries, Principal component analysis, Radio frequency, Yttrium},
}

@inproceedings{caron_emerging_2021,
	title = {Emerging Properties in Self-Supervised Vision Transformers},
	eventtitle = {Proceedings of the {IEEE}/{CVF} International Conference on Computer Vision},
	%pages = {9650--9660},
	author = {Caron, Mathilde and Touvron, Hugo and Misra, Ishan and Jégou, Hervé and Mairal, Julien and Bojanowski, Piotr and Joulin, Armand},
	% urldate = {2026-02-10},
	date = {2021},
	langid = {english},
}

@online{noauthor_visocore_nodate,
	title = {{VISOCORE}®: Secure document verification},
	url = {https://www.bundesdruckerei-gmbh.de/en/solutions/visocore},
	shorttitle = {{VISOCORE}®},
    author = {Bundesdruckerei},
	urldate = {2026-02-10},
    year = {2026},
	langid = {english},
}

@collection{angel_forensic_2020,
	location = {Boca Raton},
	title = {Forensic Document Examination in the 21st Century},
	isbn = {978-1-003-11222-8},
	doi = {10.4324/9780367853587},
	pagetotal = {336},
	publisher = {{CRC} Press},
	editor = {Angel, Miriam and Kelly, Jan Seaman},
	date = {2020-12-21},
}

@inproceedings{sevim_document_2022,
	location = {Cham},
	title = {Document Image Classification with Vision Transformers},
	isbn = {978-3-031-01984-5},
	doi = {10.1007/978-3-031-01984-5_6},
	pages = {68--81},
	booktitle = {Electrical and Computer Engineering},
	publisher = {Springer International Publishing},
	author = {Sevim, Semih and Omurca, Sevinç İlhan and Ekinci, Ekin},
	editor = {Seyman, Muhammet Nuri},
	date = {2022},
	langid = {english},
}
\end{document}